\documentclass[runningheads]{llncs}
\usepackage{graphicx}
\usepackage{amsmath,amssymb} 
\usepackage{color}
\usepackage{xspace}
\usepackage{gensymb}
\usepackage{multirow}
\usepackage{makecell}
\usepackage[caption=false]{subfig}
\usepackage[width=122mm,left=12mm,paperwidth=146mm,height=193mm,top=12mm,paperheight=217mm]{geometry}
\usepackage[activate={true,nocompatibility},final,tracking=true,kerning=true,spacing=true,factor=1100,stretch=10,shrink=10]{microtype}
\begin{document}
\pagestyle{headings}
\mainmatter

\title{Volumetric performance capture from minimal camera viewpoints} 

\titlerunning{Volumetric performance capture from minimal camera viewpoints}

\authorrunning{ A. Gilbert, M.Volino, J. Collomosse \& A. Hilton}

\author{Andrew Gilbert$^{1}$,Marco Volino$^{1}$, John Collomosse$^{1,2}$,Adrian Hilton$^{1}$}


\institute{$^{1}$Centre for Vision Speech and Signal Processing,\\
	University of Surrey\\
	$^{2}$Creative Intelligence Lab, \\
	Adobe Research\\
}

\newcommand{\etal}{{\em et~al.}\xspace}
\newcommand{\eg}{e.\,g.\xspace}
\newcommand{\ie}{i.\,e.\xspace}
\newcommand{\squeezeup}{\vspace{-4mm}}

\maketitle

\begin{abstract}

We present a convolutional autoencoder that enables high fidelity volumetric reconstructions of human performance to be captured from multi-view video comprising only a small set of camera views.  Our method yields similar end-to-end reconstruction error to that of a probabilistic visual hull computed using significantly more (double or more) viewpoints. We use a deep prior implicitly learned by the autoencoder trained over a dataset of view-ablated multi-view video footage of a wide range of subjects and actions.  This opens up the possibility of high-end volumetric performance capture in on-set and prosumer scenarios where time or cost prohibit a high witness camera count.

\end{abstract}
\squeezeup
\squeezeup
\squeezeup
\squeezeup
\squeezeup
\begin{figure}[htb]
\centering
\includegraphics[width=1\linewidth]{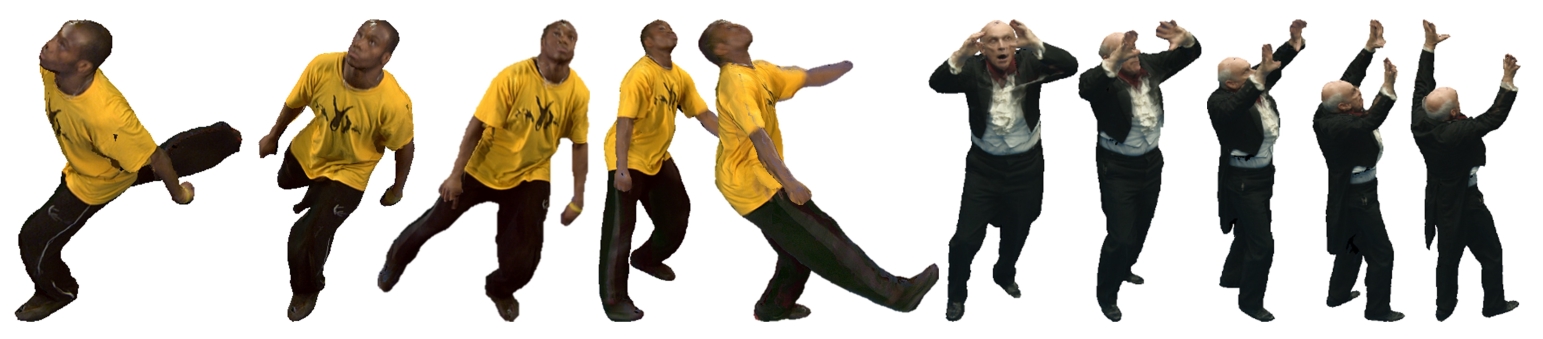}
\caption{Two high fidelity character models (JP, Magician) where 3D geometry was fully reconstructed using only two wide-baseline camera views via our proposed method.}
\label{fig:Motivation}
\end{figure}
\squeezeup
\squeezeup
\squeezeup
\section{Introduction}

Image based model reconstruction from multi-view video acquisition is enabling new forms of content production across the creative industries.  In particular, the capture of human performance in three dimensions (3D) enables rendering from an arbitrary viewpoint (free-viewpoint video rendering - FVVR) \cite{starck2009FVVR,Tsminaki2014,Casas2014EG} and photo-realistic replay within immersive VR/AR experiences. Commercial studios now operate for the capture of volumetric (``holographic") performance capture \eg implementations of at Mixed Reality Capture Studios (San Francisco, London)~\cite{collet2015MSFVV} and Intel Studios (Los Angeles) both utilising over 100 camera views of a $\sim 2.5$m$^3$ capture volume.  Whilst able to reconstruct detailed 3D models of performance, such configurations do not scale to on-set deployments where practical constraints limit the number of deployable witness cameras (\eg due to cost or rigging overheads).
The contribution of this paper is to explore whether a deeply learned prior can be incorporated into volumetric reconstruction to minimise the number of views required at acquisition.  Specifically, we investigate for the first time whether convolutional autoencoder architectures, commonly applied to visual content for de-noising and up-scaling (super-resolution),  may be adapted to enhance the fidelity of volumetric reconstructions derived from just a few wide-baseline camera viewpoints. We describe a symmetric autoencoder with 3D convolutional stages capable of refining a probabilistic visual hull (PVH) \cite{Grauman2003} \ie voxel occupancy data derived from a small set of views. Hallucinating a PVH of approximately equal fidelity to that obtainable from the same performance captured with significantly greater (double or more) camera viewpoints (Fig.~\ref{fig:Motivation}).  This extends the space of use scenarios for volumetric capture to stages with low camera counts, prosumer scenarios where cost similarly limits the number of available camera views, or settings where volumetric capture is not possible due to restrictions on camera placement and cost such as sports events~\cite{guillemaut2011joint}.


\section{Related Work}

Volumetric performance capture pipelines typically fuse imagery from multiple wide baseline viewpoints \cite{starck2009FVVR,casas2014rwvc} equispaced around the capture volume.  Initially, an estimate of volume occupancy is obtained by fusing silhouettes across views to yield a volumetric \cite{Laurentini1994} or polyhedral \cite{Franco2003} ``visual hull" of the performer. Stereo-matching and volume optimisation subsequently fuse appearance data to refine the volume estimate ultimately yielding a  textured mesh model \cite{Casas2014EG,volino2014bmvc}.  In the case of video, a 4D alignment step is applied to conform 3D mesh topology over time \cite{Budd2013}. Reconstruction error can be mitigated by temporally propagating error through a soft \ie probabilistic visual hull (PVH) \cite{Grauman2003} estimate. Or where practical by increasing the number of camera views since view sparsity limits the ability to resolve fine volume detail leading to the introduction of phantom volumes. 
Shape refinement and hole filling has been explored with a LSTM and 3D convolutional model~\cite{han2017high} for objects. 3D ShapeNets by Wu~\cite{wu20153dShapeNets}, learnt the distribution
of 3D objects across arbitrary poses and was able to discover hierarchical compositional part representation automatically for object recognition and shape completion while Sharma learnt the shape distribution of objects to enhance corrupted 3D shapes~\cite{sharma2016vconv}

Our work is inspired by contemporary super-resolution (SR) algorithms that apply learned priors to enhance visual detail in images. Classical approaches to image restoration and SR combine multiple data sources (\eg multiple images obtained at sub-pixel misalignments \cite{Fattal2007}, 
fusing these within a regularisation constraint \eg total variation \cite{tvexample}.  SR has been applied also to volumetric data in microscopy \cite{Abrahamsson2017} via depth of field, and multi-spectral sensing data \cite{Atalay2017} via sparse coding.  
Most recently, deep learning has been applied in the form of convolutional neural network (CNN) autoencoders for image \cite{Xie2012,Wang2015} and video-upscaling \cite{Shi2016}.  Symmetric autoencoders effectively learn an image transformation between clean and synthetically noisy images \cite{Jain2008} and are effective at noise reduction \eg due to image compression. Similarly, Dong \cite{Dong2016} trained end-to-end networks to learn image up-scaling.

Whilst we share the high-level goal of learning deep models for detail enhancement, our work differs from prior work including deep autoencoders in several respects.  We are dealing with volumetric (PVH) data and seek not to up-scale (increase resolution) as in SR, but instead, enhance detail within a constant-sized voxel grid to simulate the benefit of having additional viewpoints available during the formation of the PVH.  This motivates the exploration of alternative (3D) convolutional architectures and training methodologies. 

\section{Minimal Camera Volumetric Reconstruction}

\begin{figure}[t!]
\centering
\includegraphics[width=1\linewidth]{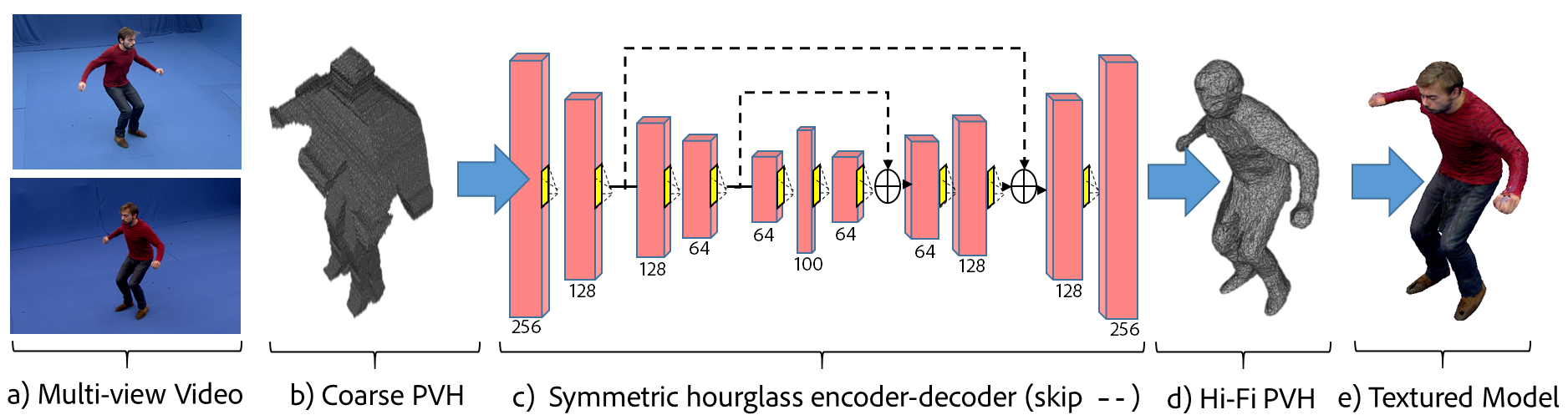}
\caption{Overview and autoencoder architecture.  A coarse PVH (b) captured using minimal camera views (a) is encoded into a latent representation via 3D convolutional and full-connected layers (c).  The decoder uses the latent representation to synthesise an output PVH of identical size but improved fidelity (d) which is subsequently meshed and textured to yield the performance capture model; meshing/texturing (e) is not a contribution of this paper.  The encoder-decoder is optimised during training using exemplar PVH pairs of the coarse and Hi-Fi PVH volumes.}
\squeezeup
\label{fig:overview}
\end{figure}

The goal of our method is to learn a generative model for high fidelity 3D volume reconstruction given a low number of wide baseline camera views. We first describe the convolutional autoencoder architecture used to learn this model using a training set of sub-volume pairs sampled from full volumetric reconstructions (PVHs) of performance obtained using differing camera counts (Sec.~\ref{sec:autoenc}). By using a PVH we are able to process wide baseline views, that would cause failure for a correspondence based method.  Our process for refining the PVH echos the stages employed in traditional image de-noising.  First, a pre-processing step (adapted from \cite{Grauman2003})  reconstructs a coarse  PVH using a limited number of cameras. This low quality result will contain phantom limbs and blocky false positive voxels (Fig.~\ref{fig:overview}b).  Next, a latent feature representation of the PVH (akin to the low-fidelity image in traditional pipelines) is deeply encoded via a series of convolution layers. We then perform non-linear mapping decoding the latent feature space to a high fidelity PVH (akin to the high-fidelity image). The reconstruction is performed in a piece-wise fashion using densely overlapping sub-volumes, This mitigates the instabilities and memory constraints of training and inference on a network with a large receptive (volumetric) field (Sec.~\ref{sec:patch}).  The high fidelity PVH is then meshed and textured with appearance data from the camera views yielding a video-realistic character model (Sec.~\ref{sec:meshing}).  Note that the final stage {\em is not a contribution of this paper}, rather we demonstrate the benefits of the PVH refinement using the method of Casas \etal \cite{Casas2014EG} but any textured meshing pipeline could be substituted as a post-process.

\subsection{Volumetric Autoencoder}
\label{sec:autoenc}
We wish to learn a deep representation given input tensor $\mathbf{V_L} \in \mathbb{R}^{X \times Y \times Z \times 1}$, where the single channel encodes the probability of volume occupancy $p(X,Y,Z)$ derived from a PVH obtained using a low camera count (eq.\ref{eq:pvh4}).  We wish to train a deep representation to solve the prediction problem $\mathbf{V_H} = \mathcal{F}(\mathbf{V_L})$ for similarly encoded tensor $\mathbf{V_H} \in \mathbb{R}^{X \times Y \times Z \times 1}$ derived from a higher fidelity PVH of identical dimension obtained using a  higher camera count.  Function $\mathcal{F}$ is learned using a CNN specifically a convolutional autoencoder consisting of successive three-dimensional (3D) alternate convolutional filtering operations and down- or up-sampling with non linear activation layers.  Fig.~\ref{fig:overview} illustrates our architecture which has symmetric structure with skip connections bridging hourglass encoder-decoder stages, the full network parameters are:

%

\begin{table}[h!]
\squeezeup
\small
\begin{tabular}{l}
$n_e$ =[64,64,128,128,256] \\
$n_d$ = [256,128,128,64,64] \\
$k_e$ = [3,3,3,3,3] \\
$k_d$= [3,3,3,3,3] \\
$k_s$ = [0,1,0,1,0] \\
NumEpoch = 10 

\end{tabular}
\label{tbl:netparams}
\squeezeup
\end{table}

\noindent where $k[i]$ indicates the kernel size and $n[i]$ is the number of filters at layer $i$ for the encoder ($e$) and decoder ($d$) parameters respectively. The location of the two skip connections are indicated by $s$ and link two groups of convolutional layers to their corresponding mirrored up-convolutional layer. The passed convolutional feature maps are summed to the up-convolutional feature maps element-wise and passed to the next layer after rectification.  The central fully-connected layer encodes the 100-D latent representation.


Learning the end-to-end mapping from blocky volumes generated from a small number of camera viewpoints to cleaner high fidelity volumes, as if made by a greater number of camera viewpoints, requires estimation of the weights $\phi$ in $\mathcal{F}$ represented by the convolutional and deconvolutional kernels. Specifically, given a collection of $N$ training sample pairs ${x^i, z^i}$, where $x^i \in \mathbf{V_L}$ is an instance of a low camera count volume and $z^i \in \mathbf{V_H}$ is the high camera count output volume provided as a groundtruth, we minimise the Mean Squared Error (MSE) at the output of the decoder across $N=X \times Y \times Z$ voxels:
\begin{eqnarray}
\mathcal{L(\phi)} = \frac{1}{N}\sum^N_{i=1} \| \mathcal{F}(x^i: \phi) -z^i \|^2_2. \label{eq:EuclidLoss}
\end{eqnarray}
To train $\mathcal{F}$ we use Adadelta~\cite{zeiler2012adadelta} an extension of Adagrad that seeks to reduce it's aggressive, radically diminishing learning rates, restricting the window of accumulated past gradients to some fixed size $w$. Given the amount of data and variation in it due to the use of patches the number of epochs required for the approach to converge is small at around 5 to 10 epochs.

\subsubsection{Skip Connections}
Deeper networks in image restoration tasks can suffer from performance degradation. Given the increased number of convolutional layers,  finer image details can be lost or corrupted, as given a compact latent feature abstraction, the recovery of all the image detail is an under-determined problem. This issue is exasperated by the need to reconstruct the additional dimension in volumetric data.  Deeper networks also often suffer from vanishing gradients and become much harder to train. In the spirit of  highway~\cite{srivastava2015trainingSkip} and deep residual networks~\cite{he2016ResNet}, we add skip connections between two corresponding convolutional and deconvolutional layers as shown in Fig.~\ref{fig:overview}.  These connections mitigate detail loss by feeding forward higher frequency content to enable up-convolutional stages to recover a sharper volume. Skip connections also benefit back-propagation to lower layers, enhancing the stability of training. Our proposed skip connections differ from that proposed in recent image restoration work \cite{srivastava2015trainingSkip,he2016ResNet} which concern only smoother optimisation. Instead, we pass the feature activation's at intervals of every two convolutional layers to their mirrored up-convolutional layers to enhance reconstruction detail.

\subsection{Volumetric Reconstruction and Sampling}
\label{sec:patch}
The low-fidelity input PVH ($\mathbf{V_L}$) is reconstructed using a variant of \cite{Grauman2003}. We assume a capture volume  observed by a limited number $C$ of camera views $c=\left[ 1,C \right]$ for which extrinsic parameters $\{R_c, {COP}_c\}$ (camera orientation and focal point) and intrinsic parameters $\{f_c, o^x_c, o^y_c\}$ (focal length, and 2D optical centre) are known, and for which soft foreground mattes are available from each camera image $I_{c}$ using background subtraction $\mathcal{BG}$.

The studio capture volume is finely decimated into voxels $\mathbf{V_L}^i=\left[\begin{array}{ccc}
v_x^i &v_y^i &v_z^i
\end{array}\right]$ for $i=\left[1, \dots, |\mathbf{V_L}|\right]$; each voxel is approximately $5 \mathrm{mm}^3$ in size. The point $(x_c,y_c)$ is the point within $I_c$ to which  $\mathbf{V_L}^i$ projects in a given view:
\begin{eqnarray}
x[\mathbf{V_L}^i]&=&\frac{f_c v_x^i}{v_z^i}+o^x_c ~~~\mathrm{and} ~~~y[\mathbf{V_L}^i]=\frac{f_c v_y^i}{v_z^i}+o^y_c,~~~\mathrm{where}\\
\left[\begin{array}{ccc}
v_x^i &v_y^i &v_z^i
\end{array}\right] &=& {COP}_c - R_c^{-1} V_L^i. \label{eq:pvh3}
\end{eqnarray}
The probability of the voxel being part of the performer in a given view $c$ is: 
\begin{eqnarray}
p(\mathbf{V_L}^i | c) = \mathcal{BG}(x[\mathbf{V_L}^i],y[\mathbf{V_L}^i]).  \label{eq:pvh1}
\end{eqnarray}
The overall likelihood of occupancy for a given voxel $p(\mathbf{V_L}^i)$ is:
\begin{eqnarray}
p(\mathbf{V_L}^i) = \prod_{i=1}^C 1/(1+e^{p(\mathbf{V_L}^i|c)}). \label{eq:pvh4}
\end{eqnarray}
We compute $p(\mathbf{V_L}^i)$ for all voxels to create the PVH for volume $\mathbf{V_L}$.  

In practice, the extent of $\mathbf{V_L}$ is limited to a sub-volume (a 3D ``patch") of the capture volume.  Patches are densely sampled to cover the capture volume, each of which is processed through $\mathcal{F}$ independently at both training and inference time.  Similar to prior image super-resolution and de-noising work \cite{Dong2016} this makes tractable the processing of large capture volumes without requiring excessively large receptive fields or up-convolutional layer counts in the CNN. In Sec.~\ref{sec:PatchEval} we evaluate the impact of differing degrees of patch overlap during dense sampling.  For efficiency we ignore any patches where $\sum_i p(\mathbf{V_L}^i)=0$.

\subsection{Meshing and texturing}
\label{sec:meshing}
Given $\mathbf{V_H}$ inferred from the network we produce a ``4D" (\ie moving 3D) performance capture. To generate the mesh for a given frame, the PVH is converted to a vertex and face based mesh using the marching-cubes algorithm. The iterative process fits vertices to the PVH output by the CNN using the marching cubes algorithm \cite{Lorensen1987} with a dynamically chosen threshold, thus producing a high-resolution triangle mesh, that is used as the geometric proxy for resampling of the scene appearance onto the texture. Without loss of generality, we texture the mesh using the approach of Casas \etal~\cite{Casas2014EG} where a virtual camera view $I_{c*}$ is synthesised in the renderer by compositing the appearance sampled from the camera views $I_{1,...,C}$ closest to that virtual viewpoint.

\begin{figure}[h!]
\centering
\includegraphics[width=1\linewidth]{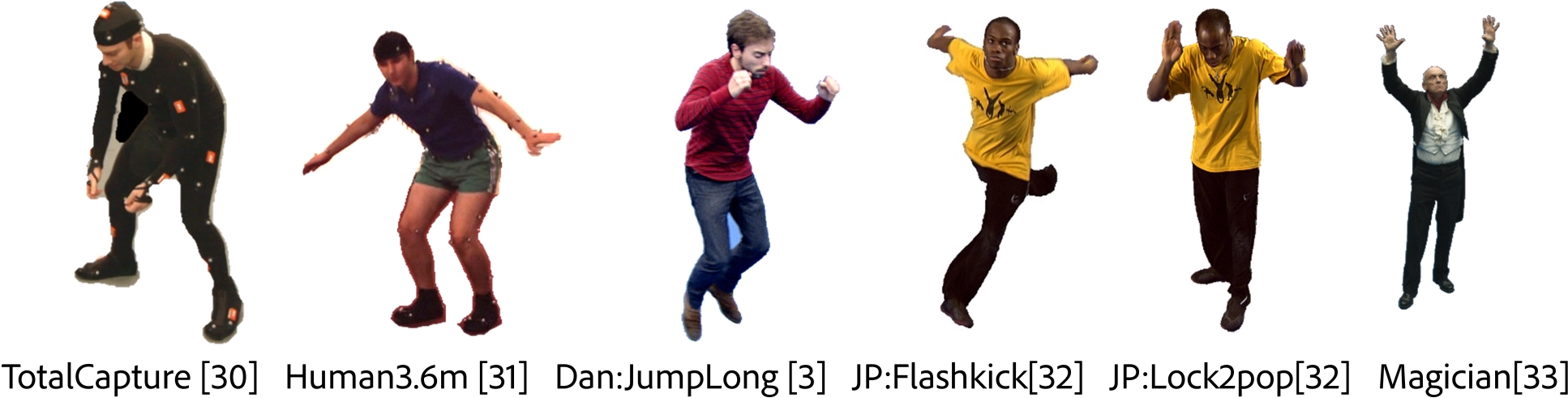}
\caption{Samples of the multi-view video datasets used to evaluate our method.}
\squeezeup
\squeezeup
\squeezeup
\label{fig:datasets}
\end{figure}

\section{Experiments and Discussion}
\label{sec:eval}

We evaluate the quantitative improvement in reconstruction accuracy, as well as the qualitative improvement in visual fidelity, due to the proposed method.  Reconstruction accuracy is evaluated using two public multi-view video datasets of human performance; \emph{TotalCapture}~\cite{trumble_total_2017} (8 camera  dataset of 5 subjects performing 4 actions with 3 repetitions at 60Hz in $360^\circ$ arrangement) and \emph{Human3.6M}~\cite{h36m_pami} (4 camera view dataset of 10 subjects performing 210 actions at 50Hz in a $360^\circ$ arrangement). Perceptual quality of textured models is evaluated using the public 4D datasets {\em Dan:JumpLong}~\cite{Casas2014EG}, {\em JP:Flashkick}~\cite{starck2007surface}, {\em JP:Lock2Pop}~\cite{starck2007surface}, and {\em Magician}~\cite{mustafa20174d}\footnote{We use the datasets released publicly at http://cvssp.org/data/cvssp3d/} (see Fig.~\ref{fig:datasets} for samples of each dataset).

\subsection{Evaluating Reconstruction Accuracy}
\label{sec:totcap}
\begin{figure}[t!]
\centering
\includegraphics[width=1\linewidth]{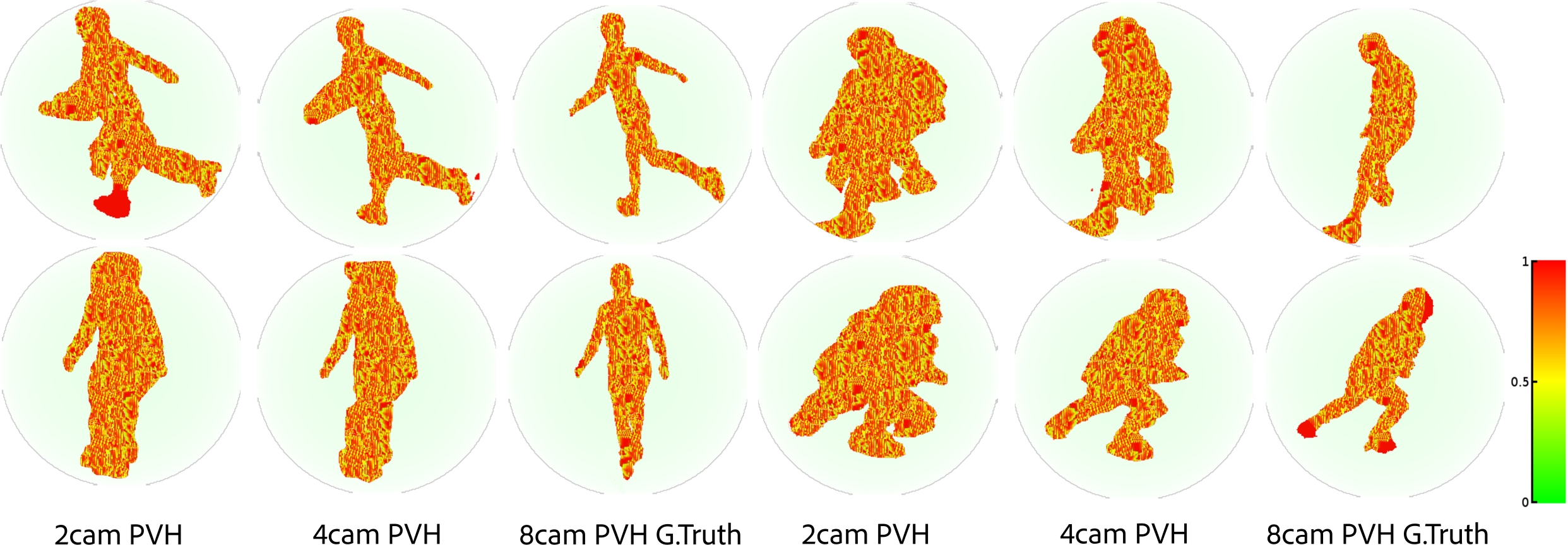}
\caption{Visualisation of raw PVH occupancy volumes estimated with C=2,4,8 views using standard method (\ie without enhancement via our approach). PVH is a probability between 0 and 1 of the subject's occupancy. This data forms the input to our auto-encoder and illustrates the phantom volumes and artefacts to contend with at $C=\{2,4\}$ versus the $C=8$ ground-truth (GT) for this dataset (\em{TotalCapture}).}
\squeezeup
\label{fig:pvhVolsBad}

\end{figure}

We study the accuracy gain due to our method by ablating the set of camera views available on {\em TotalCapture}.  The autoencoder model is trained using high fidelity PVHs obtained using all ($C=8$) views of the dataset, and corresponding low fidelity PVHs obtained using fewer views (we train for $C=2$ and $C=4$ random neighbouring views).  The model is then tested on held-out footage to determine the degree to which it can reconstruct a high fidelity PVH from the ablated set of camera views. The dataset consists of a total of four male and one female subjects each performing four diverse performances, repeated three times: \emph{ROM, Walking, Acting and Freestyle}, and each sequence lasts around 3000-5000 frames. The train and test partitions are formed wrt. to the subjects and sequences, the training consists of {\em ROM1,2,3}; {\em Walking1,3}; {\em Freestyle1,2} and {\em Acting1,2} on subjects 1,2 and 3. The test set is the performances Freestyle3 (\textbf{FS3}), Acting (\textbf{A3}) and Walking2 (\textbf{W2}) on subjects 1,2,3,4 and 5. This split allows for separate evaluation on unseen and on seen subjects but always on unseen sequences. 

The PVH is set to $z \in  \mathbb{R} ^{256 \times 256 \times 256}$.  The sub-volume ('patch') size \ie receptive field of the autoencoder ($\mathbf{V_L}$ and $\mathbf{V_H} \in \mathbb{R}^{n \times n \times n}$ is varied across $n=\{16,32,64\}$ the latter being a degenerate case where the entire volume is scaled and passed through the CNN in effect a global versus patch based filter of the volume. Patches are sampled with varying degrees of overlap; overlapping densely every 8, 16 or 32 voxels (Table~\ref{tab:QuantTC}).  The PVH at $C=8$ provides a ground-truth for comparison, whilst the $C=\{2,4\}$ input covers at most a narrow $90^\circ$ view of the scene.  Prior to refinement via the autoencoder, the ablated view PVH data exhibits phantom extremities and lacks fine-grained detail, particularly at $C=2$ (Fig.~\ref{fig:pvhVolsBad}). These crude volumes would be unsuitable for reconstruction with texture as they do not reflect the true geometry and would cause severe visual misalignments when camera texture is projected onto the model.  Applying our autoencoder method to clean up and hallucinate a volume equivalent to one produced by the unabated  $C=8$ camera viewpoints solves this issue.

\begin{figure}[htb]
\centering
\includegraphics[width=0.8\linewidth]{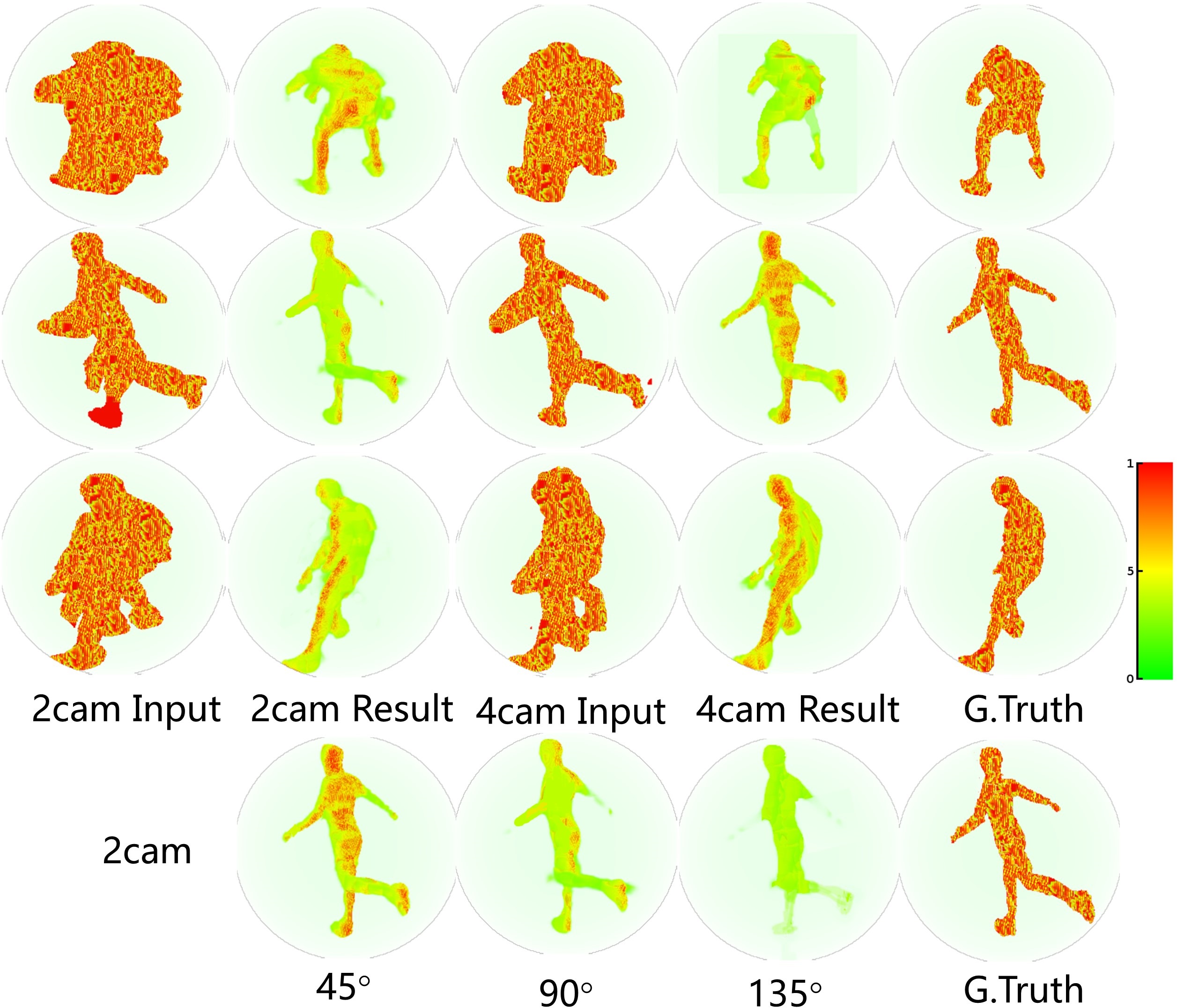}
\caption{Qualitative visual comparison of a PVH before (left) and after (right) enhancement, showing detail improvement from $C=\{2,4\}$ views {\em (TotalCapture)}. False colour volume occupancy (PVH) and groundtruth $C=8$ PVH. Bottom line indicates performance for different pairs of cameras separated by increased amounts}
\label{fig:QualTC}
\end{figure}

Table~\ref{tab:QuantTC} quantifies error between the unablated ($C=8$) and the reconstructed volumes for $C=\{2,4\}$ view PVH data, baselining these against $C=\{2,4\}$ PVH prior to enhancement via the auto-encoder (\emph{input}). To measure the performance we compute the average per-frame MSE of the probability of occupancy across each sequence. The 2 and 4 camera PVH volume prior to enhancement is also shown and our results indicate a reduction in MSE of around 4 times through our approach when 2 cameras views are used for the input and a halving of MSE for a PVH formed from 4 cameras. We observe that $C=4$ in a $180^\circ$ arc around the subject perform slightly better than $C=2$ neighbouring views in a $90^\circ$ arc. However, the performance decrease is minimal for the greatly increased operational flexibility that a 2 camera deployment provides. In all cases, MSE is more than halved (up to 34\% lower) using our refined PVH for a reduced number of views.  Using only 2 cameras, a comparable volume to that reconstructed from a full $360^\circ$ $C=8$ setup can be produced. Qualitative results of using only 2 and 4 camera viewpoint to construct the volume are shown in Figure~\ref{fig:QualTC}, where high quality reconstructions are possible despite the presence of phantom limbs and extensive false volumes in the input PVH. The bottom line includes results, from increasingly wide baseline cameras, separated by $45^\circ$, $90^\circ$, and $135^\circ$. Furthermore, the patch overlap is examined with the steps of 8,16 and 32.  When sampled at 32 voxel increments \ie without any overlap, performance is noticeably worse. This distinction between the patch overlap (16) and not (32) is visualised in Fig.~\ref{fig:64BinQual}.  In all cases, performance is slightly better when testing on seen versus unseen subjects.

\begin{table}[htb]
\centering
{
\small
\begin{tabular}{lcccccccc}
\hline
Patch   &NumCams &\multicolumn{3}{c}{SeenSubjects(S1,2,3)}        &\multicolumn{3}{c}{UnseenSubjects(S4,5)} & Mean   \\
Overlap &  C     &W2     &FS3    & A3                             &W2    &FS3    & A3                       &       \\ \hline
Input   & 2      &19.1   &28.5   &23.9                            &23.4  &27.5   &25.2                      &24.6   \\
Input   & 4      & 11.4  & 16.5  &12.5                            &12.0  & 15.2 &14.2                       & 11.6  \\ \hline
8       & 2      &5.49   & 9.98  & 6.94                           & 5.46 & 9.86  & 8.79                     & 7.75  \\
16      & 2      &5.43   & 10.03 & 6.70                           & 5.34 & 10.05 & 8.71                     & 7.71  \\
32      & 2      &6.21   & 12.75 & 8.08                           & 5.98 & 11.88 & 10.30                    & 9.20  \\  \hline
8       & 4      &5.01   & 9.07  & 6.48                           & 4.98 & 9.81  & 8.61                     & 7.33  \\
16      & 4      &5.49   & 9.56  & 6.58                           & 5.12 & 10.01 & 8.81                     & 7.60  \\
32      & 4      &5.98   & 10.02 & 7.85                           & 5.32 & 10.85 & 9.21                     & 8.28  \\ \hline
\end{tabular}
}
\caption{Quantitative performance of volumetric reconstruction on the \emph{TotalCapture} dataset using 2-4 cameras prior to our approach (Input) and after, versus unablated groundtruth using 8 cameras (error as MSE $\times 
10^{-3}$).  Patch size is $32^3$ voxels; patch overlap of 32 implies no overlap.  Our method reduces reconstruction error to 34\% of the baseline (Input) for 2 views.}
\label{tab:QuantTC}
\end{table}

\subsubsection{Cross-dataset generalisation}
\label{sec:human36}

\begin{figure}[htb]
\centering
\includegraphics[width=0.8\linewidth]{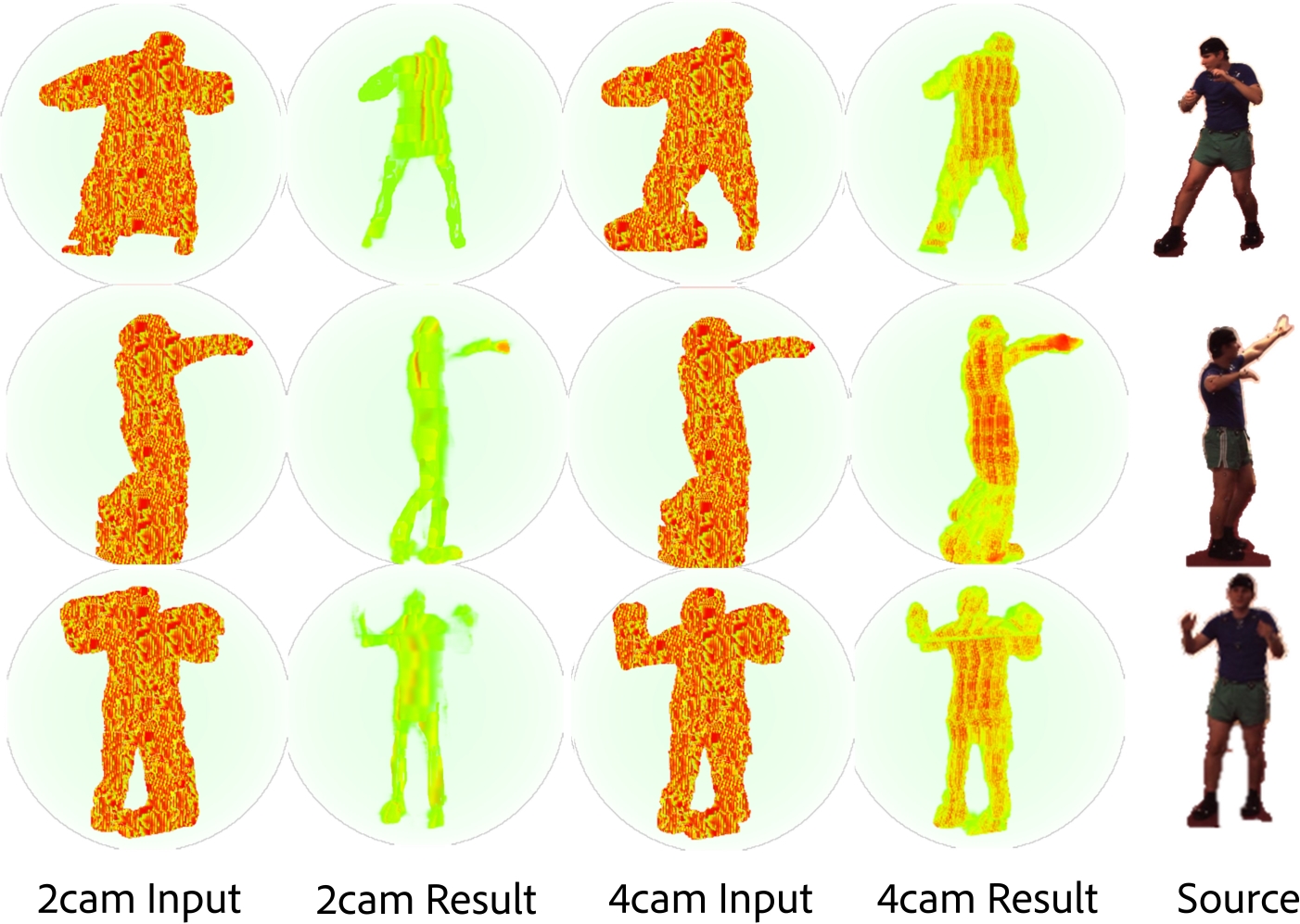}
\caption{Qualitative visual comparison of a PVH before (left) and after (right) enhancement, showing detail improvement from $C=\{2,4\}$ views {\em (Human3.6M)}. False colour volume occupancy (PVH) and source footage.}
\label{fig:human36M4camEx}
\squeezeup
\squeezeup
\end{figure}

Given that the learned model on {\em TotalCapture} can improve the fidelity of a PVH acquired with 2-4 views to approximate a PVH reconstructed from 8 views, we explore the performance of the same model on a second dataset ({\em Human3.6M}) which only has on $C=4$ views. The {\em Human3.6M} PVH models are poor quality as there are only 4 cameras at body height in four corners of a studio covering a relatively large capture area. This causes phantom parts and ghosting to occur.  Examples of the PVH  reconstructed using $C=\{2,4\}$  views on {\em Human3.6M} are shown in Fig.~\ref{fig:human36M4camEx} (red). These volumes are of poorer quality, even for 4 camera reconstructions, primarily due to the cameras being closer to the ground causing greater occlusion. However, we are able to transfer our trained CNN models for $2 \mapsto 8$ and $4 \mapsto 8$ views on {\em  TotalCapture} without any further training, to hallucinate volumes as if 8 cameras were used at acquisition. Fig.~\ref{fig:human36M4camEx} visualises the enhanced fidelity due to significantly reduced phantom volumes that would otherwise frustrate efforts to render the volume. $C=4$ result provides a more complete volume but slightly enlarged. Quantitatively, the MSE of the input PVH with $C=2$ against the \emph{groundtruth} $C=4$ PVH across the test datasets of S9 and S11 is $17.4 \times 10^{-3}$. However, after using our trained CNN model on the $C=2$ input PVH this MSE is reduced to $12.3 \times 10^{-3}$, mirroring the qualitative improvement shown in Figure~\ref{fig:human36M4camEx}.

\subsubsection{Receptive Field Size}
\label{sec:PatchEval}
\begin{figure}[htb]
\centering
\includegraphics[width=0.8\linewidth]{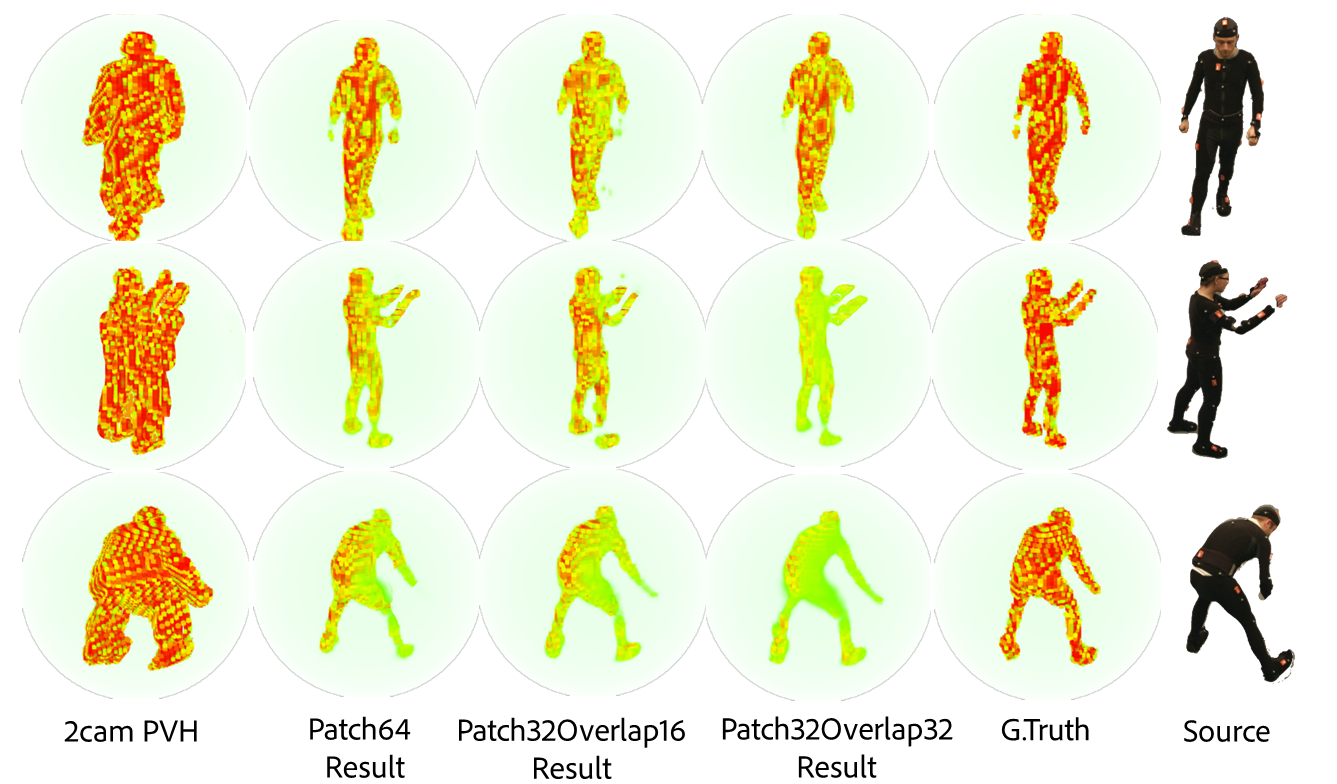}
\caption{Visual comparison accompanying quantitative data in Tbl.~\ref{tab:64BinQuant} comparing the efficacy of different patch sizes and overlaps (where a patch size of 64 implies whole volume processing).}
\label{fig:64BinQual}
\end{figure}

The use of densely sampled sub-volumes (patches) rather than global processing of the PVH is necessary for computational tractability of volumes at $ \mathbb{R}^{256 \times 256 \times 256}$, since the 3D convolutional stages greatly increase the number of network parameters and GPU memory footprint for batches during training. However, a hypothesis could be that the use of patches ignores the global context that the network could be learning about the subjects thus increasing error. Therefore we performed an experiment on the \emph{TotalCapture} dataset using the network with a modified input vector of  $z \in  \mathbb{R}^{64 \times 64 \times 64}$, therefore making each voxel around 30mm$^3$, against standard $p \in   \mathbb{R}^{32 \times 32 \times 32}$, $p \in  \mathbb{R}^{16 \times 16 \times 16}$ and $p \in   \mathbb{R}^{8 \times 8 \times 8}$ patches sampled from the same $z \in   \mathbb{R}^{64 \times 64 \times 64}$ vector, with a patch sampling overlap of 8, 16 and 32.  Quantitative results of the average MSE against the groundtruth 8 camera reconstruction volume are shown in table~\ref{tab:64BinQuant} and qualitative results are shown in Figure~\ref{fig:64BinQual}.

\begin{table}[htb]
\centering
{
\small
\begin{tabular}{lccccccccc}
\hline
Patch & Patch   &NumCams&\multicolumn{3}{c}{SeenSubjects(S1,2,3)}        &\multicolumn{3}{c}{UnseenSubjects(S4,5)} & Mean   \\
Size  & Overlap &  C     &W2     &FS3    & A3                             &W2    &FS3    & A3                       &     \\ \hline
Input & -       & 2      &20.1   &24.2   &22.3                            &23.5  &25.7   &26.8                      &23.8   \\
Input & -       & 4      &9.9    & 14.2  &13.5                            &11.8  & 14.1 &13.9                       & 12.9  \\ \hline
64    &    -    &  2     &4.34   & 6.45  & 5.78                           & 5.01 & 7.45  & 6.98                     &  6.00     \\
16    & 8       & 2      & 4.43  & 6.42  & 5.65                           & 4.99 & 7.56  & 7.23                     &  6.05     \\
16    & 16      & 2      & 5.45  & 7.03  & 6.03                           & 6.56 & 8.02  & 7.98                     &  6.85     \\ \hline
32    & 8       & 2      & 4.56  & 6.47  & 5.48                           & 5.13 & 7.98  & 6.90                     & 6.10      \\
32    & 16      & 2      & 4.42  & 6.52  & 5.63                           & 5.23 & 7.78  & 6.97                     & 6.10      \\
32    & 32      & 2      & 5.67  & 7.34  & 6.34                           & 7.02 & 8.87  & 8.03                     & 7.20      \\ \hline
\end{tabular}
}
\caption{Quantifying the effect of patch (sub-volume) size and patch overlap during dense sampling of the PVH; \emph{TotalCapture} dataset  (error as MSE $\times 10^{-3}$).}
\label{tab:64BinQuant}
\end{table}

Comparing the performance of the whole volume against patch based methods shows little change both quantitatively and qualitatively, providing that overlapping patches are utilised (therefore an overlap of 8 and 8 or 16 for $p \in  \mathbb{R}^{16 \times 16 \times 16}$ and $p \in  \mathbb{R}^{32 \times 32 \times 32}$ respectively. Therefore we can conclude that there is no requirement for global semantics to be learned as separate patches provide a measured compromise against the computational costs of training using a single global volume.  However, the benefit of using patches is that much larger PVH can be processed, as in our experiments ($256^3$ voxels). 
\subsection{4D Character Reconstruction}
\label{sec:4drecons}

We explore the efficacy of our approach as a pre-process to a state of the art 4D model reconstruction technique~\cite{Casas2014EG}.  We use three popular 4D datasets ({\em J-P}, {\em Dan}, {\em Magician}) intended to be reconstructed from a PVH derived from 8 cameras in a $360^\circ$ configuration.  We pick a subset of 2 neighbouring views at random from the set of 8, compute the low fidelity PVH from those views, and use our proposed method to enhance the fidelity of the PVH prior to running the reconstruction process~\cite{Casas2014EG} and obtaining model geometry (Sec.~\ref{sec:meshing}). The geometric proxy recovered via~\cite{Casas2014EG} is then textured using all views.  The purpose of the test is to assess the impact of any incorrect geometry on texture alignment.

The datasets all comprise a single performer indoors in a $3m^2$ capture volume. The cameras are HD resolution running at 30Hz.  Across all datasets, there are a total of 20 sequences of duration 80-3000 frames. We randomly select for test sequences: {\em Dan:JumpLong}, {\em JP:FlashKick},  {\em JP:Lock2Pop} and {\em Magician};  the remaining 16 sequences and a total of 5000 frames used as training. Given the lower number of frames available for training, the autoencoder is initially trained on {\em TotalCapture} dataset per Sec.~\ref{sec:totcap} then fine-tuned (with unfixed weights) using these 5000 frames. 
We quantify the visual fidelity of our output by rendering it from a virtual viewpoint coinciding with one of the 6 ablated viewpoints (picked randomly).  This enables a direct pixel comparison between our rendering and the original camera data for that ablated view.  As a baseline, we also compare our rendering against a baseline built using all 8 views using Casas~\cite{Casas2014EG} with identical parameters.  Each frame of test data thus yields a triplet of results for comparison; 2-view PVH, 8-view PVH, and real footage from the viewpoint.

\begin{figure}[t!]
\centering
\includegraphics[width=0.8\linewidth]{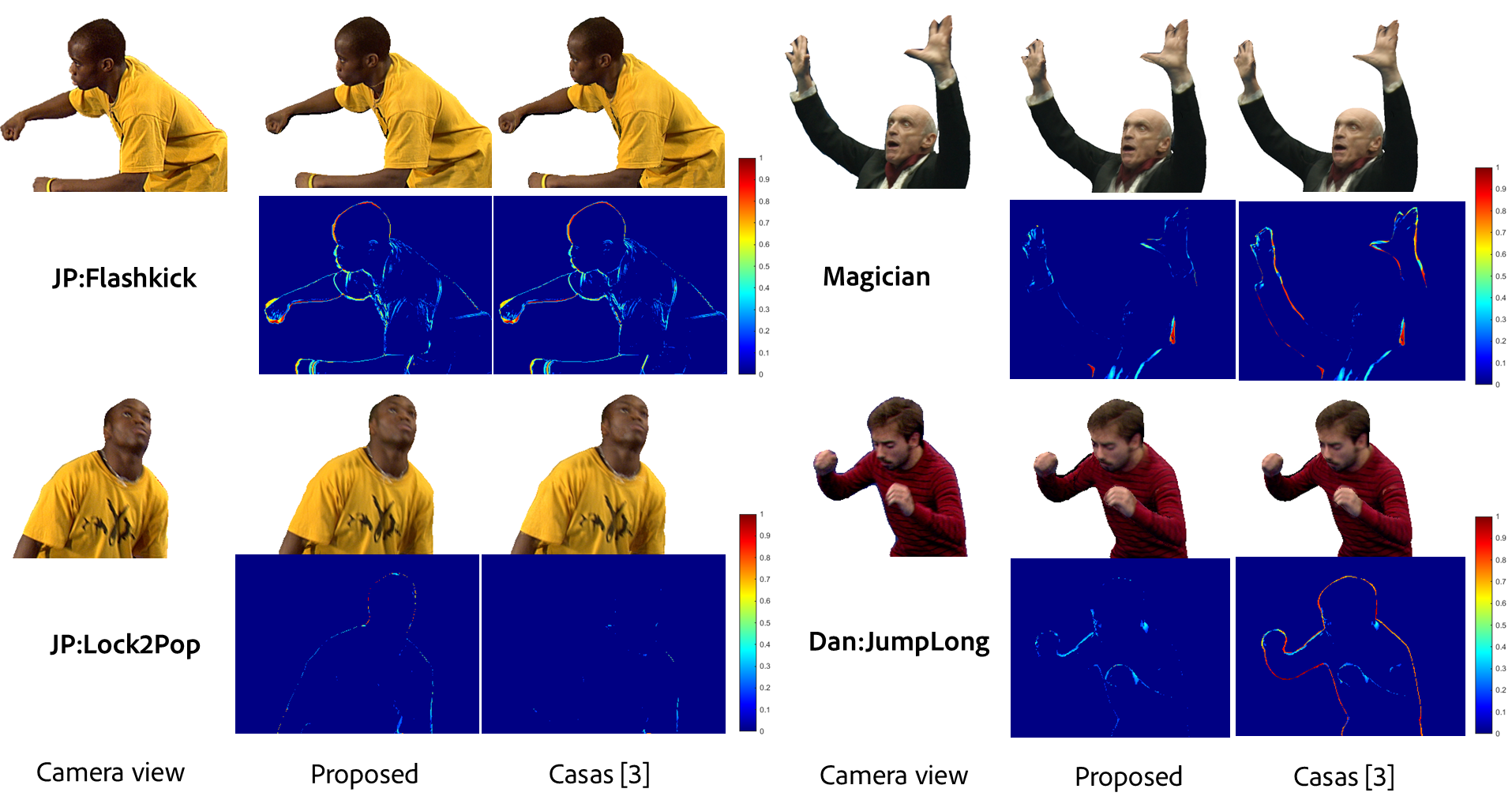}
\caption{Visual comparison of reconstructions from 2- (our) and 8-view (baseline) PVHs rendered from the viewpoint of an unused camera.  The difference images (SSIM) show only minor differences relative to the real camera footage, with 2- and 8- reconstructions near identical.  The error is quantified in Tbl.~\ref{tab:SSIM} and AMT user study (Tbl.~\ref{tab:AmazonTurk}).}
\label{fig:Viewpointcomaprsion}
\end{figure}
Fig.~\ref{fig:Viewpointcomaprsion} presents a visual comparison for a representative triplet from each of the 4 test data.  In particular, we are examining the differences in geometry which would manifest \eg via texture misalignment or spurious mesh facets that would cause texture artefacts. The results are nearly indistinguishable with only minor texture artefacts present; a high quality result considering only 2 views are used for estimate the geometry.  Table~\ref{tab:SSIM} quantifies performance using two metrics; PSNR and structural similarity (SSIM)~\cite{wang2004imageSSIM}, which closely correlates with perceptual quality. The metrics compare the 2-view and 8-view reconstructions to the camera footage which is considered to be the ground truth.
\begin{table}[!h]
\centering
{
\small
\begin{tabular}{lcccccccccc}
\hline
Method                  & \multicolumn{2}{c}{Dan}    &\multicolumn{2}{c}{JP}       &\multicolumn{2}{c}{JP}      &\multicolumn{2}{c}{Magic}   &\multicolumn{2}{c}{\multirow{2}{*}{Mean}} \\ 
                        &\multicolumn{2}{c}{JumpLong}&\multicolumn{2}{c}{flashkick}&\multicolumn{2}{c}{lock2pop}&\multicolumn{2}{c}{Magician}&\\ \hline
                        &PSNR&SSIM                   &PSNR&SSIM                    &PSNR&SSIM                   &PSNR&SSIM                   &PSNR&SSIM\\
Casas~\cite{Casas2014EG}&38.0 &0.903                 &31.8&0.893                   &32.4&0.893                  &38.1&90.4                   &35.1&0.898\\
Proposed                &37.5 &0.902                 &33.6&0.896                   &32.3&0.893                  &36.1&90.3                   &34.9&0.899\\ \hline
\end{tabular}
}
\caption{Quantifying 4D reconstruction fidelity in terms of PSNR and SSIM averaged across frames of the sequence.  We compare running \cite{Casas2014EG} over our proposed output; a PVH recovered from 2 views via the autoencoder, against a baseline reconstructed directly from an 8 view PVH.  The reconstruction errors are very similar, indicating our model correctly learns to hallucinate structure from the missing views.
}
\squeezeup
\squeezeup
\label{tab:SSIM}
\end{table}

The main sources of error between the rendered frames and the original images are found in the high frequency areas such as the face and hands, where additional vertices could provide greater detail. However, the overall reconstruction is impressive considering the poor quality of the input PVH due to the minimal camera view count.

\subsubsection{Perceptual User Study}

We conducted a study via Amazon Mechanical Turk (AMT) to compare the performance of our rendering to the 8-view baseline. A total of 500 frames sampled from the four 4D test sequences is reconstructed as above, yielding 500 image triplets.  The camera view was presented to the participant alongside the 2- and 8-view reconstructions in random order.  Participants were asked to  "identify the 3D model that is closest to the real camera image". Each result was presented 15 times, gathering in total 7763 annotations, from 343 unique users.  Tbl.~\ref{tab:AmazonTurk} reports the preferences expressed. It was our expectation that the preference would be around random at 50\%, and over the 7.8K results, yet our approach was chosen as most similar to the real camera view 50.7\% of the time. An unpaired t-test indicates the likelihood of identical preference is  $p>0.9984$. Given also the near-identical SSIM and PSNR scores we can conclude that despite only using 2 camera viewpoints our reconstructions are statistically indistinguishable from those sourced using the full 8 camera viewpoints.
\begin{table}[!h]
\centering
{
\small
\begin{tabular}{lcc}
\hline
 \\
Sequence                 & Our Approach & Casas~\cite{Casas2014EG}    \\ \hline
Dan:JumpLong                 &   43.5 \%    &       56.3\%               \\
JP:Flashkick                & 53.2 \%      &       46.7\%               \\
JP:Lock2Pop                 &  57.7\%      &       42.2\%               \\
Magician                 &  48.2\%      &       51.7\%               \\ \hline
Mean                     &  50.7        &       49.2\%               \\
Standard Deviation       &  6.15\%      &       6.11\%               \\ \hline
\end{tabular}
}
\caption{Perceptual user study (7.8k annotations). 334 AMT participants were asked to "identify the 3D model that is closest to the real camera image" and could not perceive a difference between the 2- an 8-view reconstructed models.}
\label{tab:AmazonTurk}
\end{table}

\subsection{Failure cases}
\label{sec:fail}
\begin{figure}[htb]
\centering
\includegraphics[width=0.7\linewidth]{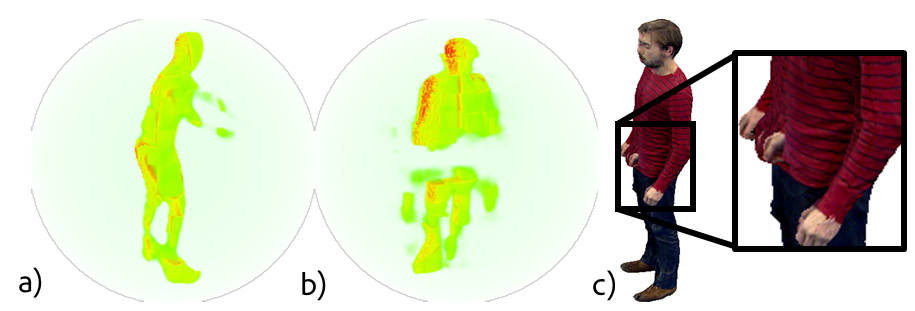}
\caption{Illustrative failure cases.  Large holes due to errors in multiple 2D mattes can cause holes in the PVH that are non-recoverable.  Texture misalignments can occur in areas of phantom geometry. Discussion in Sec.~\ref{sec:fail}.}
\label{fig:failure}
\squeezeup
\squeezeup
\squeezeup
\end{figure}

Despite the excellent performance of our approach at reconstructing view impoverished scenes, Fig.~\ref{fig:failure} highlights failure cases sometimes encountered by the proposed method. The use of the soft mattes from the 2D images to form the PVH can limit performance \eg in Fig.~\ref{fig:failure}(a) the initial coarse PVH input has a large horizontal hole and this isn't compensated for by the deeply learned prior;  in general we find the prior learns to erode phantom volumes instead of dilating existing volumes. Fig.~\ref{fig:failure}(b) illustrates that sometimes the extremities of the arms are missed, due to ambiguities in the input PVH. Finally, Fig.~\ref{fig:failure}(c), indicates a reconstruction failure due to incomplete removal of a phantom limb, caused by inaccurate geometry created from the PVH volume.

\section{Conclusion}
Volumetric performance capture from multi-view video is becoming increasingly popular in the creative industries, but reconstructing high fidelity models requires many wide-baseline views.  We have shown that high fidelity 3D models can be built with as few as a couple of views, when accompanied by a deep representation prior learned via our novel autoencoder framework.   We demonstrated that the models reconstructed via our method are quantitatively similar (Tables~\ref{tab:QuantTC},\ref{tab:64BinQuant}) and perceptually indistinguishable (AMT study, Table~\ref{tab:AmazonTurk}) from models reconstructed from considerably more camera views via existing volumetric reconstruction techniques.  An additional feature of our approach is that we are able to greatly reduce the computational cost of 4D character reconstruction. Whilst training the autoencoder takes several hours, computing the PVH and passing it through the trained network for inference of a higher fidelity volume is comfortably achievable at 25 fps on commodity GPU hardware. Furthermore, the cross-data set performance of the autoencoder appears strong without (Sec.~\ref{sec:human36}) or with minimal (Sec.~\ref{sec:4drecons}) fine-tuning.

Future work could include exploring the efficacy of our deep prior beyond the domain of human performance capture, or inference of meshes directly from a coarse PVH.  Nevertheless, we believe these findings are promising first steps toward the commoditisation of volumetric video, unlocking broader use cases for volumetric characters in immersive content.  

\section*{Acknowledgements}
The work was supported by InnovateUK via the TotalCapture project, grant agreement 102685. The work was supported in part through the donation of GPU hardware by the NVidia corporation.

\bibliographystyle{splncs}
\bibliography{egbib}

\begin{thebibliography}{10}

\bibitem{starck2009FVVR}
Starck, J., Kilner, J., Hilton, A.:
\newblock A free-viewpoint video renderer.
\newblock Journal of Graphics, GPU, and Game Tools \textbf{14}(3) (2009)
  57--72

\bibitem{Tsminaki2014}
Tsiminaki, V., Franco, J., Boyer, E.:
\newblock High resolution 3d shape texture from multiple videos.
\newblock In: Proc. Comp. Vision and Pattern Recognition (CVPR). (2014)

\bibitem{Casas2014EG}
Volino, M., Casas, D., Collomosse, J., Hilton, A.:
\newblock 4d for interactive character appearance.
\newblock In: Computer Graphics Forum (Proceedings of Eurographics 2014).
  (2014)

\bibitem{collet2015MSFVV}
Collet, A., Chuang, M., Sweeney, P., Gillett, D., Evseev, D., Calabrese, D.,
  Hoppe, H., Kirk, A., Sullivan, S.:
\newblock High-quality streamable free-viewpoint video.
\newblock ACM Transactions on Graphics (TOG) \textbf{34}(4) (2015) ~69

\bibitem{Grauman2003}
Grauman, K., Shakhnarovich, G., Darrell, T.:
\newblock A bayesian approach to image-based visual hull reconstruction.
\newblock In: Proc. CVPR. (2003)

\bibitem{guillemaut2011joint}
Guillemaut, J.Y., Hilton, A.:
\newblock Joint multi-layer segmentation and reconstruction for free-viewpoint
  video applications.
\newblock International journal of computer vision \textbf{93}(1) (2011)
  73--100

\bibitem{casas2014rwvc}
Casas, D., Huang, P., Hilton, A.:
\newblock {Surface-based Character Animation}.
\newblock In Magnor, M., Grau, O., Sorkine-Hornung, O., Theobalt, C., eds.:
  Digital Representations of the Real World: How to Capture, Model, and Render
  Visual Reality.
\newblock {CRC} Press (April 2015)  239--252

\bibitem{Laurentini1994}
Laurentini, A.:
\newblock The visual hull concept for silhouette-based image understanding.
\newblock IEEE Trans. Pattern Anal. Machine Intelligence \textbf{16}(2) (1994)

\bibitem{Franco2003}
Franco, J., Boyer, E.:
\newblock Exact polyhedral visual hulls.
\newblock In: Proc. British Machine Vision Conf. (BMVC). (2003)

\bibitem{volino2014bmvc}
Volino, M., Casas, D., Collomosse, J., Hilton, A.:
\newblock {Optimal Representation of Multiple View Video}.
\newblock In: Proceedings of the British Machine Vision Conference, BMVA Press
  (2014)

\bibitem{Budd2013}
C.Budd, Huang, P., Klaudinay, M., Hilton, A.:
\newblock Global non-rigid alignment of surface sequences.
\newblock Intl. Jnrl. Computer Vision (IJCV) \textbf{102}(1-3) (2013)  256--270

\bibitem{han2017high}
Han, X., Li, Z., Huang, H., Kalogerakis, E., Yu, Y.:
\newblock High-resolution shape completion using deep neural networks for
  global structure and local geometry inference.
\newblock Proc. Intl. Conf. Computer Vision (ICCV'17) (2017)

\bibitem{wu20153dShapeNets}
Wu, Z., Song, S., Khosla, A., Yu, F., Zhang, L., Tang, X., Xiao, J.:
\newblock 3d shapenets: A deep representation for volumetric shapes.
\newblock In: IEEE conference on computer vision and pattern recognition
  (CVPR'15). (2015)

\bibitem{sharma2016vconv}
Sharma, A., Grau, O., Fritz, M.:
\newblock Vconv-dae: Deep volumetric shape learning without object labels.
\newblock In: European Conference on Computer Vision. (2016)  236--250

\bibitem{Fattal2007}
Fattal, R.:
\newblock Image upsampling via imposed edge statistics.
\newblock In: Proc. ACM SIGGRAPH. (2007)

\bibitem{tvexample}
Rudin, L.I., Osher, S., Fatemi, E.:
\newblock Non-linear total variation based noise removal algorithms.
\newblock Physics D \textbf{60}(1-4) (1992)  259--268

\bibitem{Abrahamsson2017}
Abrahamsson, S., Blom, H., Jans, D.:
\newblock Multifocus structured illumination microscopy for fast volumetric
  super-resolution imaging.
\newblock Biomedical Optics Express \textbf{8}(9) (2017)  4135--4140

\bibitem{Atalay2017}
Aydin, V., Foroosh, H.:
\newblock Volumetric super-resolution of multispectral data.
\newblock In: Corr. arXiv:1705.05745v1. (2017)

\bibitem{Xie2012}
Xie, J., Xu, L., Chen, E.:
\newblock Image denoising and inpainting with deep neural networks.
\newblock In: Proc. Neural Inf. Processing Systems (NIPS). (2012)  350--358

\bibitem{Wang2015}
Wang, Z., Liu, D., Yang, J., Han, W., Huang, T.S.:
\newblock Deep networks for image super-resolution with sparse prior.
\newblock In: Proc. Intl. Conf. Computer Vision (ICCV). (2015)  370--378

\bibitem{Shi2016}
Shi, W., Caballero, J., Huszar, F., Totz, J., Aitken, A., Bishop, R., Rueckert,
  D., Wang, Z.:
\newblock Real-time single image and video super-resolution using an efficient
  sub-pixel convolutional neural network.
\newblock In: Proc. Comp. Vision and Pattern Recognition (CVPR). (2016)

\bibitem{Jain2008}
Jain, V., Seung, H.:
\newblock Natural image denoising with convolutional networks.
\newblock In: Proc. Neural Inf. Processing Systems (NIPS). (2008)  769--776

\bibitem{Dong2016}
Dong, C., Loy, C.C., He, K., Tang, X.:
\newblock Image super-resolution using deep convolutional networks.
\newblock IEEE Trans. Pattern Anal. Machine Intelligence \textbf{38}(2) (2016)
  295--307

\bibitem{zeiler2012adadelta}
Zeiler, M.D.:
\newblock Adadelta: an adaptive learning rate method.
\newblock arXiv preprint arXiv:1212.5701 (2012)

\bibitem{srivastava2015trainingSkip}
Srivastava, R.K., Greff, K., Schmidhuber, J.:
\newblock Training very deep networks.
\newblock In: Advances in neural information processing systems. (2015)
  2377--2385

\bibitem{he2016ResNet}
He, K., Zhang, X., Ren, S., Sun, J.:
\newblock Deep residual learning for image recognition.
\newblock In: Proceedings of the IEEE conference on computer vision and pattern
  recognition. (2016)  770--778

\bibitem{Lorensen1987}
Lorensen, W., Cline, H.:
\newblock Marching cubes: A high resolution 3d surface construction algorithm.
\newblock ACM Transactions on Graphics (TOG) \textbf{21}(4) (1987)  163--169

\bibitem{trumble_total_2017}
Trumble, M., Gilbert, A., Malleson, C., Hilton, A., Collomosse, J.:
\newblock Total capture: 3d human pose estimation fusing video and inertial
  sensors.
\newblock In: Proceedings of 28th British Machine Vision Conference.  1--13

\bibitem{h36m_pami}
Ionescu, C., Papava, D., Olaru, V., Sminchisescu, C.:
\newblock Human3.6m: Large scale datasets and predictive methods for 3d human
  sensing in natural environments.
\newblock IEEE Transactions on Pattern Analysis and Machine Intelligence
  \textbf{36}(7) (jul 2014)  1325--1339

\bibitem{starck2007surface}
Starck, J., Hilton, A.:
\newblock Surface capture for performance-based animation.
\newblock IEEE computer graphics and applications \textbf{27}(3) (2007)

\bibitem{mustafa20174d}
Mustafa, A., Volino, M., Guillemaut, J.Y., Hilton, A.:
\newblock 4d temporally coherent light-field video.
\newblock 3DV 2017 Proceedings (2017)

\bibitem{wang2004imageSSIM}
Wang, Z., Bovik, A.C., Sheikh, H.R., Simoncelli, E.P.:
\newblock Image quality assessment: from error visibility to structural
  similarity.
\newblock IEEE Tran. Image Processing (TIP) \textbf{13}(4) (2004)  600--612

\end{thebibliography}
\end{document}